\documentclass{llncs}

\usepackage[a4paper,includeheadfoot,top=1.65in,bottom=1.65in,left=1.73in,hcentering]{geometry}
\usepackage[pdftex]{graphicx}
\usepackage[sectionbib]{natbib}
\usepackage{caption} 
\usepackage[utf8]{inputenc}
\usepackage{t1enc}
\usepackage{tikz-dependency}
\usepackage{qtree}
\usepackage{float}
\usepackage[title]{appendix}

\usepackage{booktabs}
\usepackage{array}
\usepackage{multirow}
\usepackage{amsmath}
\usepackage{algpseudocode}
\usepackage{commath}
\usepackage{tabularx,booktabs}

\DeclareMathOperator*{\argmax}{arg\,max}

\usepackage{subcaption}
\restylefloat{figure}
\usepackage{tikz-qtree}
\usepackage[english]{babel}
\usepackage{algorithm}

\renewcommand\bibname{References}

\newif{\ifhidecomments}
\hidecommentsfalse

\ifhidecomments
    \newcommand{\botond}[1]{}
    \newcommand{\attila}[1]{}
    \newcommand{\judit}[1]{}
    \newcommand{\dorina}[1]{}
\else
    \newcommand{\botond}[1]{\textcolor{red}{[#1 ({\bf Botond})]}}
    \newcommand{\attila}[1]{\textcolor{blue}{[#1 ({\bf Attila})]}} 
    \newcommand{\judit}[1]{\textcolor{orange}{[#1 ({\bf Judit})]}} 
    \newcommand{\dorina}[1]{\textcolor{green}{[#1 ({\bf Dorina})]}}
\fi

\begin{document}

\title{Data Augmentation for Machine Translation via Dependency Subtree Swapping}
	\author{Attila Nagy\inst{1}, Dorina Petra Lakatos\inst{1, 2}, Botond Barta\inst{1, 2}, Patrick Nanys\inst{1}, Judit \'Acs\inst{2}\\
\institute{
$^1$Department of Automation and Applied Informatics\\
Budapest University of Technology and Economics\\
$^2$Institute for Computer Science and Control\\
Eötvös Loránd Research Network}
\email{attila.nagy234@gmail.com},
\email{patrick.nanys2000@gmail.com},
\email{\{dorinalakatos, botondbarta, acsjudit\}@sztaki.hu}  
}

\maketitle

\begin{abstract}
We present a generic framework for data augmentation via dependency subtree swapping that is applicable to machine translation. We extract corresponding subtrees from the dependency parse trees of the source and target sentences and swap these across bisentences to create augmented samples. We perform thorough filtering based on graph-based similarities of the dependency trees and additional heuristics to ensure that extracted subtrees correspond to the same meaning. We conduct resource-constrained experiments on 4 language pairs in both directions using the IWSLT text translation datasets and the Hunglish2 corpus. The results demonstrate consistent improvements in BLEU score over our baseline models in 3 out of 4 language pairs. Our code is available on GitHub\footnote{https://github.com/attilanagy234/syntax-augmentation-nmt}.
\end{abstract}

\section{Introduction}
Parallel data is a necessity for building performant neural machine translation (NMT) systems. For high- and medium-resource languages, millions of parallel data points enable researchers to build translation models of high quality.
However, in a resource-constrained setting, such as low-resource or domain-specific machine translation, the lack of data must be compensated with a variety of techniques to improve performance. These include training multilingual models for zero- and few-shot learning \citep{johnson2017google, sharaf2020meta}, transfer learning \citep{kocmi-bojar-2018-trivial} and a wide variety of data augmentation techniques that alter existing parallel sentences to create more training data \citep{sennrich-etal-2016-improving, zhang-zong-2016-exploiting, fadaee-etal-2017-data}.

In this paper, we present a framework for generating augmented samples with the help of syntactic information for machine translation. We benchmark our augmentation method in a low-resource setting on 4 language pairs, using low-resource datasets for English-German, English-Hebrew and English-Vietnamese and a subsample of a high-resource dataset for English-Hungarian.

\section{Related work}
Data augmentation (DA) involves a set of techniques that enhance the data used to train a machine learning model both in size and variety.
It has been widely applied in NLP for fixing class imbalance, mitigating bias, making the model more robust through adverserial examples or increasing model accuracy \citep{feng-etal-2021-survey}. 
DA methods that incorporate syntactic knowledge have been applied to a number of NLP tasks, but not particularly in the domain of machine translation. For part-of-speech tagging, \cite{sahin-steedman-2018-data} augment dependency trees by removing dependency links and moving tree fragments around the root. This dependency tree morphing method was also shown effective for dependency parsing by \cite{vania-etal-2019-systematic}. \cite{xu-etal-2016-improved} improve relation classification by making use of the directionality of relationships in a dependency tree. The idea of swapping compatible subparts of datapoints to generate augmented samples has shown performance improvements in dependency parsing \citep{dehouck-gomez-rodriguez-2020-data}, named entity recognition \citep{dai-adel-2020-analysis} and constituency parsing \citep{shi-etal-2020-role}. \cite{shi-etal-2021-substructure} introduce a generalized framework for substructure substitution, which produces new samples by swapping same-label substructures.

For machine translation, the most common augmentation method is backtranslation \citep{sennrich-etal-2016-improving}, where additional training data is obtained by translating a target-language monolingual corpus into the source language, using a baseline model trained with the originally available data.
\cite{fadaee-etal-2017-data} targets low-frequency words by generating new sentence pairs with rare words in a new context. \cite{wang-etal-2018-switchout} propose SwitchOut, a technique, which randomly replaces words in both the source and target sentences with other random words from their corresponding vocabularies. \cite{gao-etal-2019-soft} introduce Soft Contextual DA, where they augment a randomly chosen word in a sentence by its contextual mixture of multiple related words. \cite{NEURIPS2020_7221e5c8} diversify the training data by using the predictions of multiple forward and backward models and then merging them with the original dataset on which the final NMT model is trained. \cite{moussallem2019augmenting} use knowledge graphs to enhance semantic feature extraction and hence the translation of entities and terminological expressions. \cite{ijcai2020p496} create constraint-aware training data by first randomly sampling the phrases of the reference as constraints, and then packing them together into the source sentence with a separation symbol. \cite{wei-etal-2020-uncertainty} propose an uncertainty-aware semantic DA method, which explicitly captures the universal semantic information among multiple semantically-equivalent source sentences and enhances the hidden representations with this information for better translations. \cite{sanchez-cartagena-etal-2021-rethinking} present an approach, where augmented sentence pairs are generated with simple transformations and used as auxiliary tasks in a multi-task learning framework with the aim of providing new contexts where the target prefix is not informative enough to predict the next word. \cite{wei-etal-2022-learning} suggest Continuous Semantic Augmentation (CSANMT), which augments each training instance with an adjacency semantic region that could cover adequate variants of literal expression under the same meaning. A method, which uses dependency parsing for data augmentation was introduced by \cite{duan2020syntax}. They perform simple token-level manipulations, such as blanking, dropout and replacement and use the depth of tokens in the dependency tree as guidance for computing word-selection probabilities.

\section{Methodology}
\begin{figure}[!htbp]
\centerline{
\scalebox{0.95}{
    \includegraphics[width=0.99\textwidth]{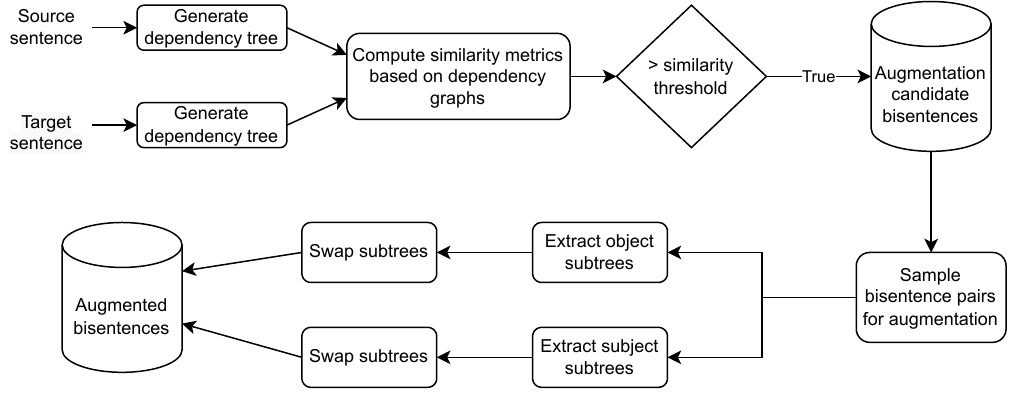}
    \caption{Overview of our data augmentation method}
    \label{fig:syntax-aug-system-overview}
}
}
\end{figure}

In this section, we introduce the data augmentation and the graph-based filtering methods used in our experiments in detail. Figure \ref{fig:syntax-aug-system-overview} gives an overview of the augmentation algorithm starting from pairs of parallel sentences or bisentences.

\subsection{Dependency subtree swapping}
\begin{figure}[!htbp]
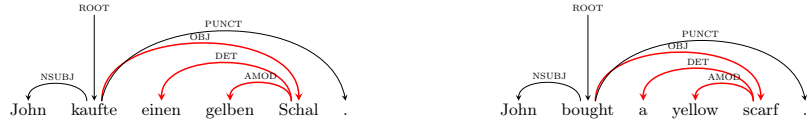

\begin{subfigure}{0.5\textwidth}
\centering
\scalebox{.70}{
    \begin{dependency}[theme = simple]
    \centering
      \begin{deptext}[column sep=1em]
          John \& kaufte \& einen \& gelben \& Schal \& . \\
      \end{deptext}
      \deproot{2}{ROOT}
      \depedge[arc angle=90]{2}{1}{NSUBJ}
      \depedge[arc angle=90, edge style = {red, thick}]{2}{5}{OBJ}
      \depedge[arc angle=90, edge style = {red, thick}]{5}{3}{DET}
      \depedge[arc angle=90, edge style = {red, thick}]{5}{4}{AMOD}
      \depedge{2}{6}{PUNCT}
    \end{dependency}
}
\end{subfigure}
\hspace*{\fill}
\begin{subfigure}{0.5\textwidth}
\centering
\scalebox{.70}{
    \begin{dependency}[theme = simple]
    \centering
      \begin{deptext}[column sep=1em]
          John \& bought \& a \& yellow \& scarf \& . \\
      \end{deptext}
      \deproot{2}{ROOT}
      \depedge[arc angle=90]{2}{1}{NSUBJ}
      \depedge[arc angle=90, edge style = {red, thick}]{2}{5}{OBJ}
      \depedge[arc angle=90, edge style = {red, thick}]{5}{3}{DET}
      \depedge[arc angle=90, edge style = {red, thick}]{5}{4}{AMOD}
      \depedge{2}{6}{PUNCT}
    \end{dependency}
}
\end{subfigure}
\caption{Example of a object substructures in the dependency trees of a German-English translation pair. The corresponding subtrees are highlighted in red.}
\label{fig:dep-tree-example}
\end{figure}

\begin{figure}[!htbp]
\centerline{
\scalebox{0.99}{
    \includegraphics[width=0.99\textwidth]{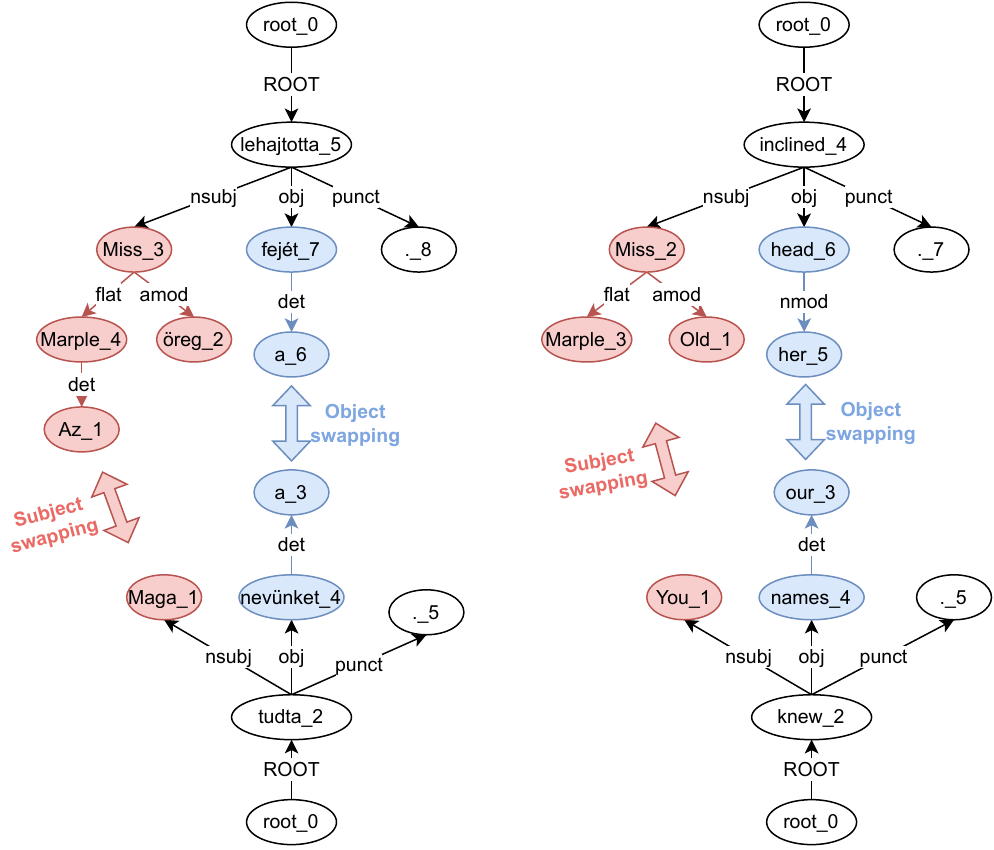}
    \caption{Two kinds of augmentation techniques: object and subject subtree swapping.}
    \label{fig:swapping}
}
}
\end{figure}

We propose a simple, yet effective data augmentation method for machine translation based on dependency trees. The intuition behind our approach is that elements of a translation pair - although they are of a different language - show syntactic similarities. The central idea is to extract corresponding syntactic substructures in the source and target sentences and swap these across bisentences to produce augmented datapoints. One clear advantage of this method compared to many other non-model based data augmentation methods for NMT, is that this approach alters the source and target sentences simultaneously, so it has a better chance of maintaining parallelism in the augmented samples. Depending on the language and the phrasing, associated substructures could have different composition, although we hypothesize that subjects and objects have a similar enough representation across many language-pairs to be used for this kind of augmentation. We consider object and subject subtrees in the dependency trees that correspond to OBJ and NSUBJ edges defined in the Universal Dependencies \citep{nivre-etal-2020-universal}. Figure~\ref{fig:dep-tree-example} shows an example of the object subtrees in a German-English translation pair. Figure~\ref{fig:swapping} explains the technique of object and subject subtree swapping through an example bisentence pair.

Applying the method in a more generic setting to more complex syntactic substructures is possible. However, as the performance of the algorithm is dependent on both the translation quality and the quality of the dependency parser, finding correspondance between more complex subtrees could be hard and that may result in injecting too much noise into the training data. Nevertheless, this is an interesting future direction of research, which should be explored.

\subsection{Generating augmentation candidates}
Our prior experiments have shown that although the number of samples that we can generate grows exponentially with respect to the size of the training data, the generated bisentences can be very noisy \citep{nagy2022syntax}. 
In this section, we discuss the rules and heuristics that we implemented to ensure that the generated bisentences are high quality.
\subsubsection{Filtering heuristics}
As the augmentation method is based on swapping certain subtrees, it is only possible to sample from a subset of the training data, where the dependency trees contain the required edges. We consider a bisentence eligible for augmentation, if the dependency trees of both the source and target sentences have at least one NSUBJ and one OBJ edges. We also constrain that they cannot contain more than one of each dependency edge type. In order to further reduce noise, we exclude bisentences, where the root of the selected subtrees of the source and target sentences do not belong to the same part-of-speech tag. We also constrain, that every subtree eligible for swapping must contain either a noun or a proper noun.
\subsubsection{Graph-based sampling}
As the method is sensitive to both the translation quality and the performance of the dependency parser, we need to ensure that the selected subtrees capture the same meaning in both the source and target sentences. We hypothesize that if the syntactic composition of the subtrees of the source and target sentences are significantly different, the subtrees likely have a different meaning. To study the syntactic similarities of the dependency trees, we explore two graph-based methods: Graph Edit Distance (GED) \citep{6313167} and Edge Mapping (EM).

Graph Edit Distance is a generalization of the Levenshtein distance \citep{levenshtein1966binary} for graphs. The GED value represents the minimal cost of graph edit operations, which is required to transform the first graph into the second one. The operations we used for nodes and edges are insertion, deletion and substitution. A deletion or an insertion costs 1, while the substitution costs 2. To compare these numbers concerning the size of the graphs, we normalized the edit distance with the following equations:
\begin{equation}
\begin{aligned}
\label{eq:ged}
    d_{\text{max}} &= 2 * |V_1| - 1 + 2 * |V_2| - 1 \\
    \text{sim}(G_1,G_2) &= \dfrac{d_{\text{max}} - \text{GED}(G_1,G_2)}{d_{\text{max}}}
\end{aligned}
\end{equation}
where $d_{\text{max}}$ corresponds to the maximum distance between 2 graphs, when all of the edges and nodes are deleted from $G_1$ and every node and edge in $G_2$ is inserted. $V_1$ and $V_2$ are the set of vertices in $G_1$ and $G_2$ respectively. The $\text{sim}(G_1, G_2)$ is a similarity measure between $G_1$ and $G_2$, and its value is between 0 and 1, 1 if the two graphs are isomorphic.

\begin{figure}[!htbp]
\centerline{
\scalebox{0.99}{
    \includegraphics[width=0.99\textwidth]{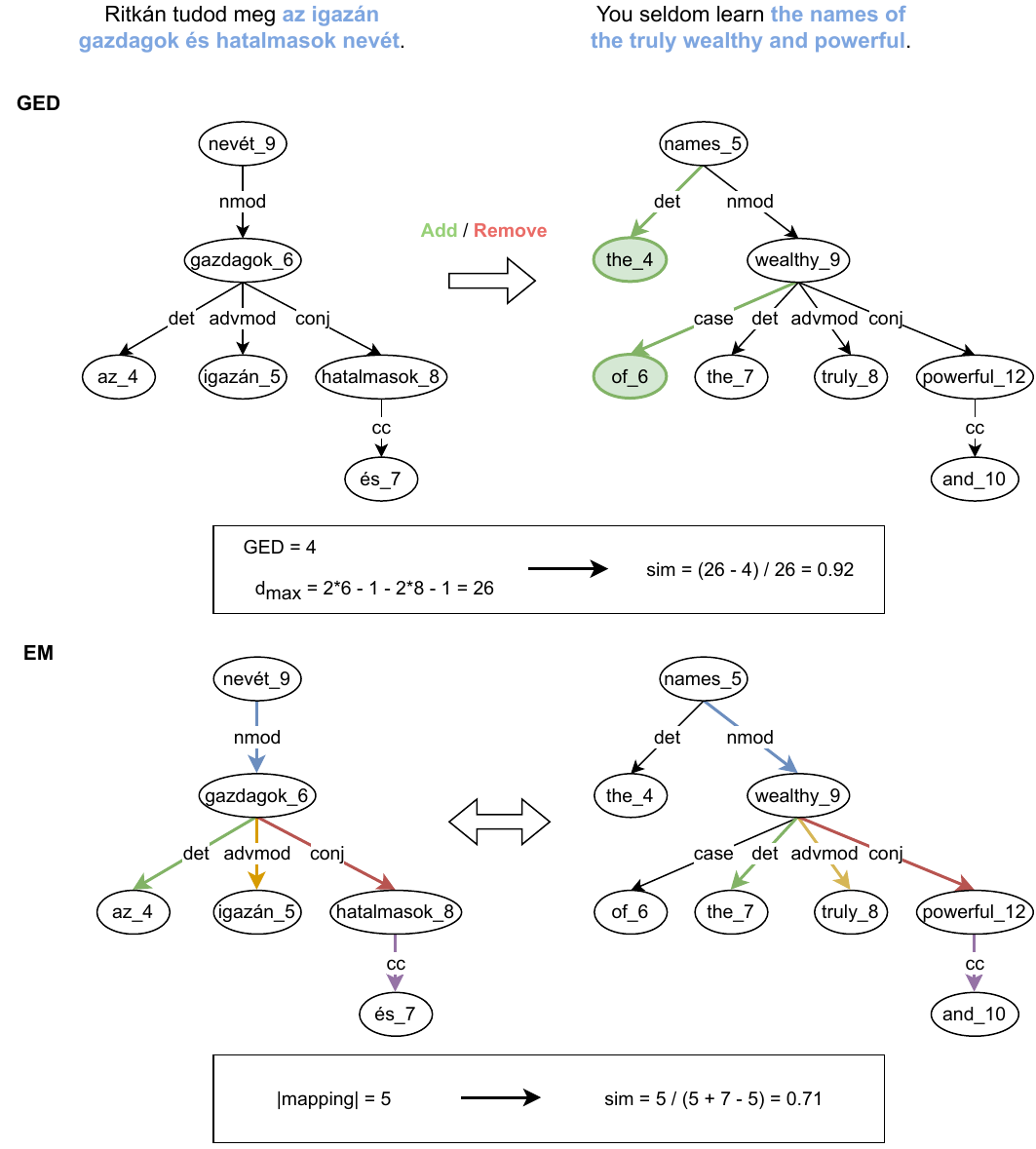}
    \caption{The two kinds of graph similarity measure of a bisentence's object subtree. For the GED method the added and removed vertices/edges are colored green and red respectively. In the case of the EM measure, the colors of the edges represent the mapping.}
    \label{fig:graph_example}
}
}
\end{figure}

The Edge Mapping (EM) algorithm is based on a general graph similarity algorithm \citep{graph_sim}. A modified version of this is described in Algorithm \ref{alg:mapping}. The $\text{score}(e_1, e_2)$ represents how many nodes are common between the two edges (2 at max). $\text{route\_sim}(e_1,e_2)$ is the Levenshtein distance between the $\text{root}-e_1$ and $\text{root}-e_2$ routes, where every route is defined by the part-of-speech tag of the visited nodes. The algorithm gives a mapping between the edges of the graphs, which we consider as the intersection. This way we can calculate a Jaccard index between the edges, which gives us another similarity measure:
\begin{equation}
\label{eq:jaccard}
    J(G_1,G_2) = \dfrac{|m|}{|E_1| + |E_2| - |m|}
\end{equation}
where $m$ is the mapping, $E_1$ and $E_2$ are the set of edges in $G_1$ and $G_2$ respectively. Figure~\ref{fig:graph_example} compares the two kinds of similarity measures through the object subtree of a sentence.

\begin{algorithm}[t]
\caption{Edge mapping}
\label{alg:mapping}
\begin{algorithmic}
\Require $G_1(V_1,E_1),G_2(V_2,E_2)$
\State $m \gets \{\}$
\ForAll {$e_1 \in E_1$}
\State $\text{cands} \gets \{e_2 \mid e_2 \in E_2, e_2 \notin m, e_1=e_2\}$
\If{$\text{cands}$ is empty}
    \State \textbf{continue}
\EndIf
\State $\text{cands} \gets \argmax\limits_{c \in \text{cands}} \text{score}(e_1, c)$
\State $\text{cands} \gets \argmax\limits_{c \in \text{cands}} \text{route\_sim}(e_1, c)$
\State $m[e_1] \gets \text{cands}[0]$
\EndFor
\State \Return m
\end{algorithmic}
\end{algorithm}

\section{Experiments}
We conduct experiments on 4 language pairs in both directions: English--German, English--Hebrew, English--Vietnamese and English--Hungarian. We trained baseline models for all language pairs, which we benchmark against models trained using additional augmented data.

\subsection{Datasets}
Following other works in the field \citep{sanchez-cartagena-etal-2021-rethinking, gao-etal-2019-soft, guo-etal-2020-sequence, wang-etal-2018-switchout}, we use datasets from the text translation track of IWSLT. The IWSLT datasets are transcripts of Ted and TedX talks and are regularly used for benchmarking data augmentation methods for NMT in a low-resource setting. For English-German and English-Hebrew, we use the IWSLT 2014 \citep{cettolo-etal-2014-report} text translation track for training and we use the \textit{tst2013} and \textit{tst2014} datasets for development and test sets respectively. For English-Vietnamese, we select the IWSLT 2015 \citep{cettolo-etal-2015-iwslt} text translation track for training, whereas the \textit{tst2012} and \textit{tst2013} datasets are used for development and test sets respectively. For English-Hungarian, we perform experiments on the Hunglish2 corpus \citep{varga2007parallel}, which is a sentence-aligned dataset. It was constructed by scraping and aligning bilingual text in several domains, such as literature, movie subtitles, software documentation and legal text. To simulate a low-resource experiment, we created a stratified subsample of 250k sentences based on the domains and created a train-dev-test splits from this, also with stratified sampling.

We apply the same preprocessing steps for all language pairs. We remove sentences longer than 32 tokens. We also remove sentences where the difference of the source- and target-side token counts are more than 7 and their ratio is more than 1.2. We also remove leading and trailing quotation marks and dashes. We perform language detection on both the source and target sentences using \textit{fastText} \citep{joulin2016fasttext} and remove datapoints whenever the languages mismatch with what is expected. We normalize punctuations with \textit{sacremoses}\footnote{https://github.com/alvations/sacremoses}. The statistics of the the preprocessed datasets can be seen in Table \ref{table:dataset-stats}.

\begin{table}[!htbp]
  \centering
    \begin{tabular}{ l r r r }
    \toprule
    \textbf{Language pair} & \textbf{train} & \textbf{dev} & \textbf{test} \\
    \midrule
    En-De & 127,506 & 993 & 1,305 \\
    En-He & 132,105 & 1,382 & 962 \\
    En-Vi & 89,188 & 1,553 & 1,268 \\
    En-Hu & 212,500 & 37,500 & 21,700 \\
    \bottomrule
    \end{tabular}
  \caption{Number of bisentences in the train/dev/test sets for each language pair.}
  \label{table:dataset-stats}
\end{table}

\subsection{Training details}
We use the same Transformer-based encoder-decoder architecture \citep{vaswani2017attention} for every experiment. The hyperparameters used in each experiment can be seen in Table \ref{table:base-parameters}.
All of the models were trained on a single A100 GPU for 4 hours using the openNMT framework \citep{klein-etal-2017-opennmt}. We used early stopping to avoid overfitting on the training set based on the perplexity of the validation set. For tokenization, part-of-speech tagging and dependency parsing, we use HuSpacy \citep{HuSpaCy:2021} for Hungarian and Stanza \citep{qi2020stanza} for the rest of the languages.

\begin{table}[!htbp]
\centering
\begin{tabular}{lr|lr}
\toprule
\textbf{Parameter}          & \textbf{Value}    & \textbf{Parameter}    & \textbf{Value}       \\ \midrule
batch type & tokens & batch size & 12288\\
accumulation count & 4 & average decay & 0.0005 \\
train steps & 150000 & valid steps & 5000 \\
early stopping & 4 & early stopping criteria & ppl \\
optimizer & adam & learning rate & 2 \\
warmup steps & 8000 & decay method & noam \\
adam beta2 & 0.998 & max grad norm & 2 \\
label smoothing & 0.1 & param init & 0 \\
param init glorot & true & normalization & tokens \\
max generator batches & 32 & encoder layers & 8 \\
decoder layers & 8 & heads & 16 \\
RNN size & 1024 & word vector size & 1024 \\
Transformer FF & 2096 & dropout steps & 0\\
dropout & 0.1 & attention dropout & 0.1\\
share embeddings & true & position encoding & true \\
\bottomrule
\end{tabular}
\caption{Hyperparameters of the models.}
\label{table:base-parameters}
\end{table}

\begin{figure}[!htbp]
\centerline{
\scalebox{0.99}{
    \includegraphics[width=0.99\textwidth]{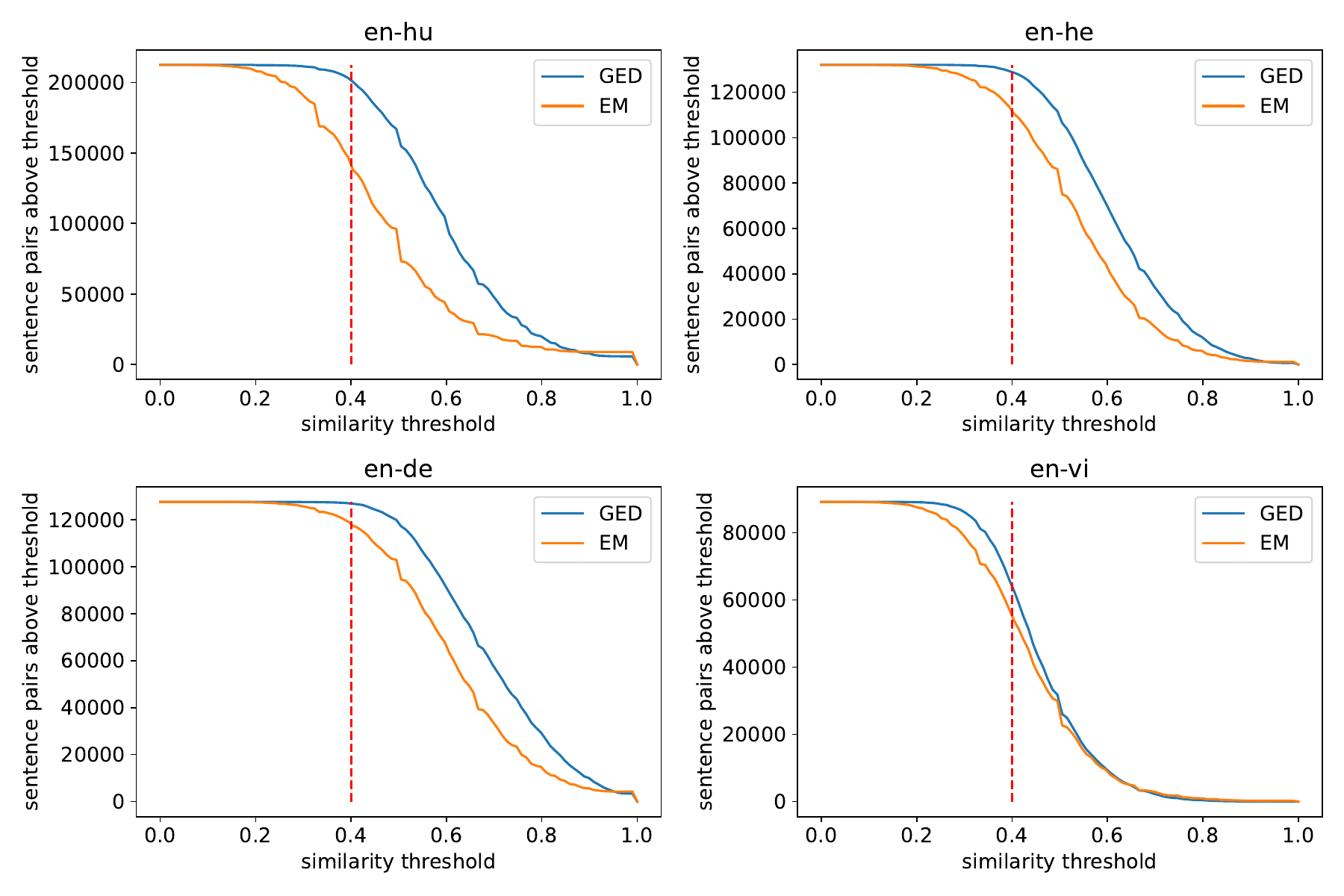}
    \caption{The number of sentence pairs with similarity above a given threshold 
    for both the GED and EM methods computed between the dependency parse trees of the source and target sentences.}
    \label{fig:graph-similarities}
}
}
\end{figure}

For each experiment, we generate augmented data and shuffle it with the training sets before training. We explore data generation with different augmentation configurations, including the graph similarity metrics (GED and EM), the subtree type that is swapped (Subj, Obj) and different augmentation ratios ($0.5$, $2$, $3$). For the graph similarity filtering, we use $0.4$ as a generic threshold for all language pairs. We found during qualitative analysis of augmented data points that this value yields high quality and diverse results. Selecting a higher threshold value does not necessarily mean better quality augmentations, because in practice, a very high similarity often means that the subtrees are small and this way we cannot guarantee with high confidence that the subtrees correspond to the same meaning. The number of sentence pairs with similarity above a given threshold between the source and target sentences are shown for all language-pairs in Figure \ref{fig:graph-similarities}.

\section{Results}

\begin{figure}
\centering\includegraphics[width=0.99\textwidth]{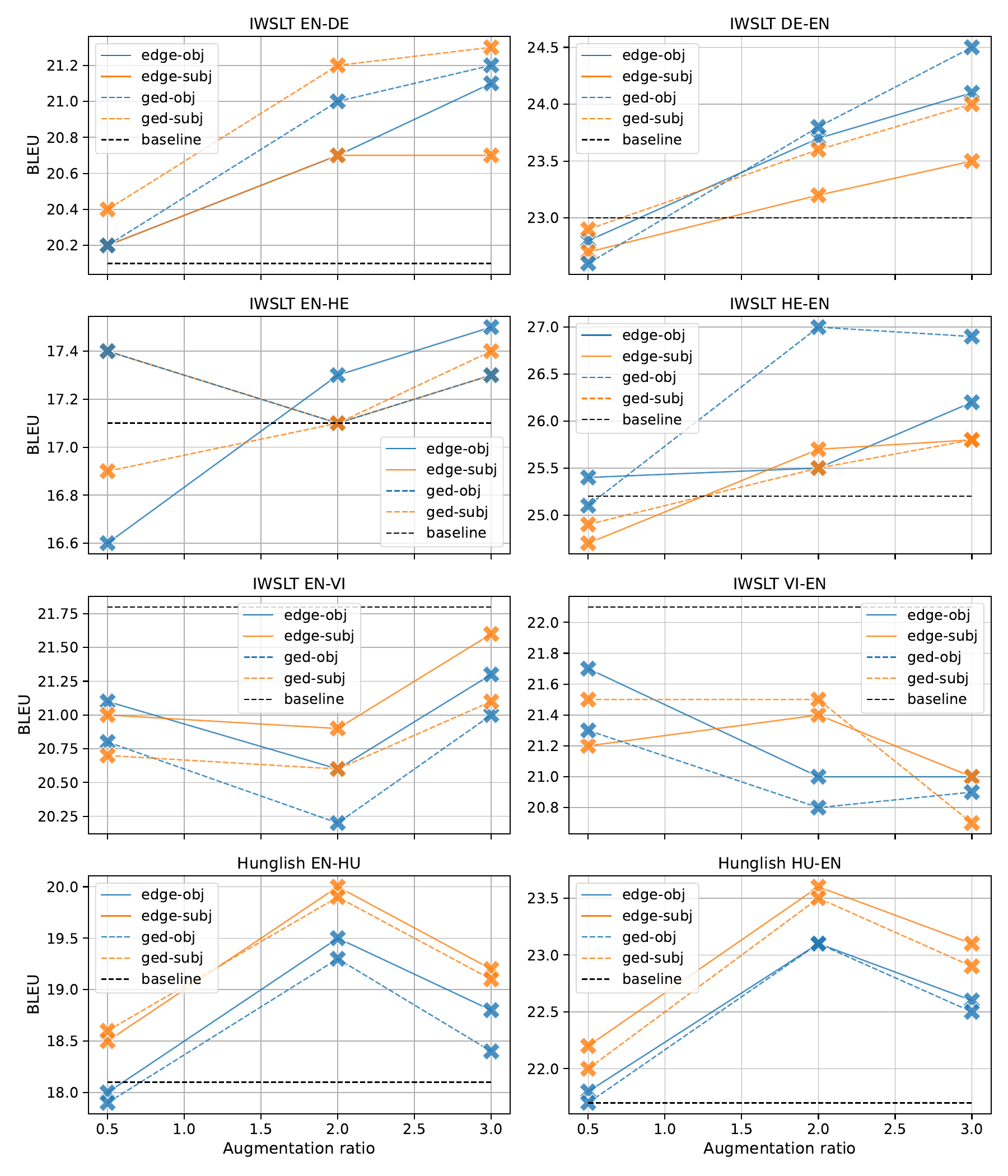}
\caption{Test BLEU scores for each augmentation method and augmentation ratio. We indicate the baseline results with a horizontal black dashed line.}
\label{fig:results}
\end{figure}

We measure the effectiveness of the proposed method with BLEU scores of NMT models trained with and without augmented data. 
Figure~\ref{fig:results} shows the BLEU scores for each dataset and each augmentation setup.
It is clear that there is no \emph{one size fits all}, but with the exception of Vietnamese, at least one configuration is always better than the baseline with a non-negligible margin.
Unfortunately the Vietnamese dependency parser is not very high quality, which explains the subpar performance after our augmentation.
We find an upward trend in the other three language pairs, in other words, more augmented data seems to help, although the exact augmentation ratio needs to be determined separately for each dataset.
Table~\ref{table:aug-results} reports the BLEU scores of the baseline models and the best performing models' hyperparameters and results.

\begin{table}[!htbp]
  \centering
    \begin{tabular}{ l | r | c |  c | r | r }
    \hline
    \textbf{Language pair} & \textbf{Baseline} & \textbf{Graph sampling} & \textbf{Aug type} & \textbf{Aug ratio} & \textbf{Aug BLEU}\\
    \hline
    En-De & 20.1 & GED & Subj & 3 & 21.3 \\
    De-En & 23.0 & GED & Obj & 3 & 24.5 \\
    En-He & 17.1 & EM & Obj & 3 & 17.5 \\
    He-En & 25.2 & GED & Obj & 2 & 27.0 \\
    En-Vi & 21.8 & EM & Subj & 3 & 21.6 \\
    Vi-En & 22.1 & EM & Obj & 0.5 & 21.7 \\
    En-Hu & 18.1 & EM & Subj & 2 & 20.0 \\
    Hu-En & 21.7 & EM & Subj & 2 & 23.6 \\
    \hline
    \end{tabular}
  \caption{Augmentation parameters of the best performing models for each language pair.}
  \label{table:aug-results}
\end{table}

In general we observed that the augmentation method starts to improve the performance of the models considerably if the augmentation ratio is 2 or higher. This means augmenting a dataset of size $X$ with at least $2X$ amount of new samples.
The choice of augmentation type (Object or Subject) seems to depend on the language.
Subject is usually better, particularly for Hungarian, but it is less obvious in other languages.
The same can be said about the filter type (GED or EM).
One trend, we see is that the same augmentation type and ratio usually yields similar results regardless of the filter type.
This suggests that the filter methods we use are somewhat robust.

The augmentation methods did not improve the scores for English-Vietnamese in neither directions. We suspect that this is because Stanza has relatively poor performance on tokenization and dependency parsing for Vietnamese\footnote{https://stanfordnlp.github.io/stanza/performance.html} . It would be interesting to reproduce the experiments with a better NLP pipeline for Vietnamese and compare the results.

\section{Conclusion}
We presented an augmentation method for machine translation based on dependency subtree swapping. 
We applied two types of graph distance based filtering to discard low quality augmentation.
We demonstrated the effectiveness of our method by performing resource-constrained experiments on 4 language pairs in both directions. 
With the exception of English--Vietnamese, we see consistent improvement over the non-augmented baselines in English--German, English--Hebrew and English--Hungarian.

\section*{Acknowledgements}
The authors would like to thank András Kornai and Csaba Oravecz for discussions on the augmentation methodology.

\newpage
%
\renewcommand\bibname{References}
\bibliographystyle{splncsnat_en}
\bibliography{mszny}

\end{document}